\documentclass[journal,,twocolumn,transmag]{IEEEtran}
\usepackage{caption}
\usepackage{times}
\usepackage{epsfig}
\usepackage{graphicx}
\usepackage{amsmath}
\usepackage{amssymb}
\usepackage{multirow}
\usepackage{epstopdf}
\usepackage{color}

\ifCLASSINFOpdf
\else
\fi
\begin{document}
%

\title{Multiple Discrimination and Pairwise CNN for View-based 3D Object Retrieval}
\author{\IEEEauthorblockN{
Z. Gao,
K.X Xue,
S.H Wan, Senior Member, IEEE}
}
%
\maketitle
\begin{abstract}
Abstract--With the rapid development and wide application of computer, camera device, network and hardware technology, 3D object (or model) retrieval has attracted widespread attention and it has become a hot research topic in the computer vision domain. Deep learning features already available in 3D object retrieval
have been proven to be better than the retrieval performance of hand-crafted features. However, most existing networks do not take into account the impact of multi-view image selection on network training, and the use of contrastive loss alone only forcing the same-class samples to be as close as possible. In this work, a novel solution named Multi-view Discrimination and Pairwise CNN (\textbf{MDPCNN}) for 3D object retrieval is proposed to tackle these issues. It can simultaneously input of multiple batches and multiple views by adding the Slice layer and the Concat layer. Furthermore, a highly discriminative network is obtained by training samples that are not easy to be classified by clustering. Lastly, we deploy the contrastive-center loss and contrastive loss as the optimization objective that has better intra-class compactness and inter-class separability. Large-scale experiments show that the proposed \textbf{MDPCNN} can achieve a significant performance over the state-of-the-art algorithms in 3D object retrieval. 

\footnote{This work was supported in part by the National Natural Science Foundation of China (No.61872270, No.61572357), National Key R\&D Program of China (No.2019YFBB1404700), Jinan 20 projects in universities (No.2018GXRC014). Tianjin Municipal Natural Science Foundation (No.14JCZDJC31700, No.13JCQNJC0040).

Z. Gao is with Qilu University of Technology (Shandong Academy of Sciences), Shandong Computer Science Center (National Supercomputer Center in Jinan), Shandong Artifical Intelligence Institute, Jinan, 250014, P.R China. 

K.X Xue and Z. Gao are with Key Laboratory of Computer Vision and System, Ministry of Education, Tianjin University of Technology, Tianjin, 300384, China.


S.H Wan (Corresponding author) is with School of Information and Safety Engineering, Zhongnan University of Economics and Law, Wuhan 430073, China (Email:shaohua.wan@ieee.org).


}

\end{abstract}


\maketitle

\IEEEdisplaynontitleabstractindextext

%
\IEEEpeerreviewmaketitle

\section{Introduction}
%
%
%
%
3D object retrieval plays an extremely important role in the computer vision domain and has a very extensive application prospect.  The existing 3D object retrieval methods can be divided into two categories: three-dimensional (3D) models based retrieval method and two-dimensional (2D) views based retrieval method. In the retrieval algorithms based on model structure, low-level features (\cite{Yasseen2016View,Osada2002Shape}) such as model texture, the geometry shape, or high-level features (\cite{Mademlis20093D,LENG2009MATE}) such as model topology, are often employed to achieve 3D object retrieval. However, these methods are very demanding for the integrity of the model. While the view-based retrieval algorithm is implemented by obtaining 2D images of a 3D object. Therefore the retrieval of 3D objects is converted to retrieve 2D images of those objects. This method is also widely adopted by developers, and researchers have proposed some visual features for view-based 3D object retrieval, e.g., Zernike moments (\cite{Khotanzad1990Invariant}), Elevation Descriptors (EDs), Light-Field Descriptors (LFDs) (\cite{Chen2003On}), Bag of Visual Features (BoVF) (\cite{Ohbuchi2009Scale}), Histogram of Orientation Gradient (HOG) (\cite{Dalal2005Histograms,Shih2007A}),  Compact Multi-view Descriptors (CMVDs) (\cite{Ohbuchi2008Salient}), and Multi-view and Multivariate Gaussian descriptor (MMG) (\cite{Gao2017Multi}). Accordingly, 3D object retrieval algorithms have emerged one after another, for instance, Adaptive Views Clustering (AVC) (\cite{Ansary2006A}), Nearest Neighbor (NN) (\cite{Steinbach2000A}), Weighted Bipartite Graph Matching (WBGM) (\cite{Gao20113D}), Camera Constraint-Free View (CCFV) (\cite{Gao2012Camera}), Hausdorff Distance (HAUS) (\cite{Gao20113D}), and Class-statistics matching method with pair constraint (CSPC) (\cite{Gao2016A,Gao2019Cognitive}). However, as we all know, the selection of different features or retrieval algorithms will affect the performance of 3D object retrieval (\cite{Liu2018View}). Since these hand-crafted features may have certain emphasis and unilateralism when there are different illuminations, different shooting angles, or very small differences of different object categories, thus, the retrieval performances of these features are very unstable, and it also limits the improvement of the retrieval algorithms.

In view of the fact that many researchers have considered the use of convolutional neural networks to extract more robust deep learning features to recognize tasks. Convolutional neural network (CNN) has distinctive advantages in speech recognition and image recognition, and different CNNs  (\cite{L1998Gradient,Wang2020Cerebral,Krizhevsky2012ImageNet,Wang2019Cerebral,Simonyan2014Very,Szegedy2015Going,He2015Deep,Wang2020Alcoholism}) have been proposed such as LeNet-5 (\cite{L1998Gradient}),  AlexNet  (\cite{Krizhevsky2012ImageNet}), VggNet (\cite{Simonyan2014Very}), GoogleNet (\cite{Szegedy2015Going}) and ResNet (\cite{He2015Deep}). Although the features acquired by CNNs are obviously superior to other traditional hand-crafted features, it is still necessary for retrieval tasks to do further processing after finishing the deep feature extraction, in other words, it is not end-to-end network architecture. This separation of tasks causes inadequate training and it also takes a long time to finish the whole retrieval. Therefore, an end-to-end network architecture called Siamese network  (\cite{Chopra2005Learning,Norouzi2012Hamming}) arises at the historic moment. Meanwhile, various networks  (\cite{Wang2015Sketch,Su2016Multi,Gao2018GroupPair}) have been upgraded on the basis of their own fields. However, in these algorithms, the selection of multi-view images in network training is not well discussed where the random selection scheme is often employed. Moreover,  the network discrimination also needs to be improved in which the same-class samples are forced to be as close as possible, but the inter-class samples are ignored.

In order to solve the issues of 3D object retrieval better, we put forward a new method named Multi-view Discrimination and Pairwise CNN (\textbf{MDPCNN}), which is a discriminating multi-batch and multi-view pairwise convolution neural network. The Slice layer and the Concat layer in the network are designed to realize the simultaneous input of multiple batches and multiple views. These two layers are only used to split and integrate data, and the time consuming is negligible because no computation is required. Due to the input of pairwise multi-view images, a large number of training samples will be generated in the actual network training process. Thereupon we train \textbf{MDPCNN} by selecting samples which are not easily classified so, thus, a highly discriminating network can be obtained. Finally, the contrastive and the contrastive-center losses are proposed to work together on the network to achieve the purpose of intra-class compactness and inter-class separability. A large scale experiments show that \textbf{MDPCNN} has acquired excellent results, making the network more stable and efficient.
  The main contributions of this paper are summarized as follows.
\begin{itemize}
\item We add the Slice layer and the Concat layer to the network so that we can simultaneously enter multiple batches and multiple views. This can speed up the convergence of the network and improve the convergence accuracy of the network.
\item We use clustering scheme to select samples which are the most difficult to classify, to make the network more distinguishable.
\item We adopt the contrastive and the contrastive-center losses as the optimization target to improve the discrimination of the network, which will promote its performance improvement.
\item Large-scale experiments on four benchmarks of 3D objects demonstrate that \textbf{MDPCNN} has a significant advantage over the state-of-the-art 3D object retrieval algorithms.
\end{itemize}

The rest of the paper is organized as follows. We first discuss some related works in \textbf{Related Work}, and then we elaborate our proposed \textbf{MDPCNN} algorithm. Moreover, we present experiments to evaluate our method in \textbf{Experiments}. Finally, the conclusions of this paper is given in \textbf{Conclusion}.

 \section{Related Work}

So far, different CNN architectures have been proposed for 3D object retrieval, where different network architectures and different loss functions are designed. Moreover, in our work, it also belongs to multi-view pairwise CNNs. Thus, the related work will be presented from two aspects: (1) Network structure; (2) Loss function. We will describe these two parts in the following sections separately.

\begin{itemize}
\item In general, the 3D object retrieval process includes feature extraction and object retrieval algorithms. In order to integrate these two steps to get end-to-end network architecture, the Siamese network architecture (\cite{Chopra2005Learning,Norouzi2012Hamming}) was proposed where a similarity measure from the data is learned, and then it is utilized to compare and match samples of the new unknown category. The main advantage of this network is that it weakens the label, makes the network scalable and classifies those categories that have not been trained. Moreover, this algorithm is also applicable to some small scale datasets, which can increase the size of the entire dataset in disguise, so that the dataset with relatively small data volume, can be trained by the deep network to have a good effect. Compared with direct training with 3D data, the maturity and speed of 2D image convolution network have great advantages. MVCNN (\cite{Su2016Multi}) exploits 2D rendering maps of 3D objects from different perspectives as the original training data, and uses the classic and mature 2D image convolution network for training. It combines multi-view features (\cite{Gao2019Adaptive}) with a CNN architecture called "view-pooling layer" to create a single and concise 3D shape descriptor. Using a 2D image to understand the 3D object is often better than using a 3D object directly. The feature of directly using the separated a 2D projection image can also receive good results, but the feature of integrating all the features of a 2D projection image to describe a 3D object is much better. However, because MVCNN is not an end-to-end network architecture for 3D object retrieval, and it takes some time to extract deep learning features and then identify or classify them. Besides that, MVCNN inputs multiple views where the location of the batch-size is occupied, thus, the batch size cannot be directly set in the network, which will affect the convergence speed and its accuracy. According to the different advantages of the Siamese network and MVCNN, Gao et al. (\cite{Gao2018GroupPair}) gave an end-to-end double-chain network structure called GPCNN. GPCNN takes the form of pairwise view as an input unit, and matches the similarity of two view groups to get the similarity of two objects and then realizes the retrieval process. It implements an end-to-end retrieval network based on the network structure of the Siamese network and MVCNN combined with contrastive loss. GPCNN can solve the problem of the small scale of original samples, effectively mining the correlation information between multiple views. Nevertheless, GPCNN only randomly selects images to form the group pairs without considering the impact of different views on the training network. Additionally, it uses contrastive loss only to bring the distance of the same class samples closer, and does not deal with different class samples. In addition, a hierarchical view-group-shape architecture, such as i.e., from the view level, the group level and the shape level, is proposed in GVCNN (\cite{Feng2018GVCNN}) where the view-pooling scheme is utilized to generate a group level description, and the shape level descriptor is generated by combining all group level descriptors according to their discriminative weights. Besides, hyper-graph neural networks (HGNN) framework (\cite{Feng2019Hypergraph}) is proposed to study the feature representation where the high-order data correlation is encoded in a hyper-graph structure. Although the intrinsic hierarchical correlation and discriminability among views are well exploited in GVCNN and HGNN, it does not belong to the Siamese network, thus, the large scale training samples are needed. When the training samples are limited, its performance is not satisfying.

\item In the model learning, the loss function is often utilized to estimate the degree of inconsistency between the true value and prediction value, whose value is a real and non-negative. Ordinarily, the smaller the loss function is, the better the robustness of the model. The Softmax function is often used in CNN networks, and when the input of the Softmax, the multinomial logistic loss of it is calculated. By this way, a more numerically stable gradient can be obtained, for example, in GVCNN  (\cite{Su2016Multi}), the hierarchical and complex network architecture is designed, and multi-view information is simultaneously inputted the network, but only the Softmax function is employed in it. Although the Softmax function is often utilized in single chain network, but it is not suitable for the Siamese network. Thus, the loss function adopted by the Siamese network and GPCNN, is a contrastive loss, which can effectively manage the relationship of pairwise data in the Siamese neural network (\cite{Hadsell2006Dimensionality}). This loss function was primarily used in dimensionality reduction. That is to say, after dimensionality reduction (feature extraction), the two samples are still similar in feature space, while the original dissimilar samples are still not similar in feature space. Contrastive loss shortens sample distance if the label is 1 and expands sample distance if the label is 0. This loss function weakens the label of the sample itself and does not take into account the factor of the categories in which the sample belongs. Softmax is the most widely used classification loss, which maps the output of multiple neurons to the (0, 1) interval to get the probability distribution. In many cases, the inter-class spacing is even larger than the intra-class spacing. We expect that features are not only separable from each other, but also have well discriminative. The combination of simple Softmax and center loss (\cite{Wen2016A}) can train features that are cohesive. Center loss wants the square of the distance from the center of the feature for each sample in a batch to be smaller, that is, the distance within the class is as small as possible. Softmax can be assumed to be responsible for increasing the inter-class distance, and center-loss for reducing the intra-class distance, so that the learned feature discrimination will be higher. Both Softmax and center losses only consider the optimization of one situation, but these two situations are very important, thus, the discrimination loss function should be designed where intra-class compactness and inter-class separability should be simultaneously assessed by penalizing the contrastive values: (1) the distances between training samples from the same category and their corresponding category centers should be as small as possible, and (2) the sum of the distances of training samples to their non-corresponding category centers should be as big as possible. Conversely, it cannot handle the relationship of pairwise data in the Siamese neural network.

\end{itemize}

\section{Proposed Method}
The problem that 3D object retrieval urgently needs to be solved is how to better describe a 3D object with multiple 2D views. Inspired by (\cite{Hadsell2006Dimensionality,Su2016Multi,Qi2017Contrastive,Gao2018GroupPair}), a novel method named \textbf{MDPCNN} is proposed to deal with 3D object retrieval. The \textbf{MDPCNN} framework for 3D object retrieval is given in Fig.1. In this network architecture, it consists of sample selection, multi-batch and loss function. In sample selection, the training samples are chosen by clustering to form many group pairs. The implementation of multiple batches and multiple views is accomplished through the Slice layer and the Concat layer. Fig.2 show how the slice layer divide the feature mapping to view pooling layer. Contrastive loss and contrastive-center loss are simultaneously employed to enhance the distinctiveness of the samples. On the basis of the slice layer, we can obtain the mapping features for each view respectively, and then compute the maximum value for each element in each mapping image across different views which is called as the element-wise operation. Finally, the max pooling is employed in sub-sampling. This procedure is called as the view-pooling. In the following sections, we will explain them in detail.

\begin{figure*}[t]
\begin{center}
\includegraphics[width=6.8in,height = 3.in]{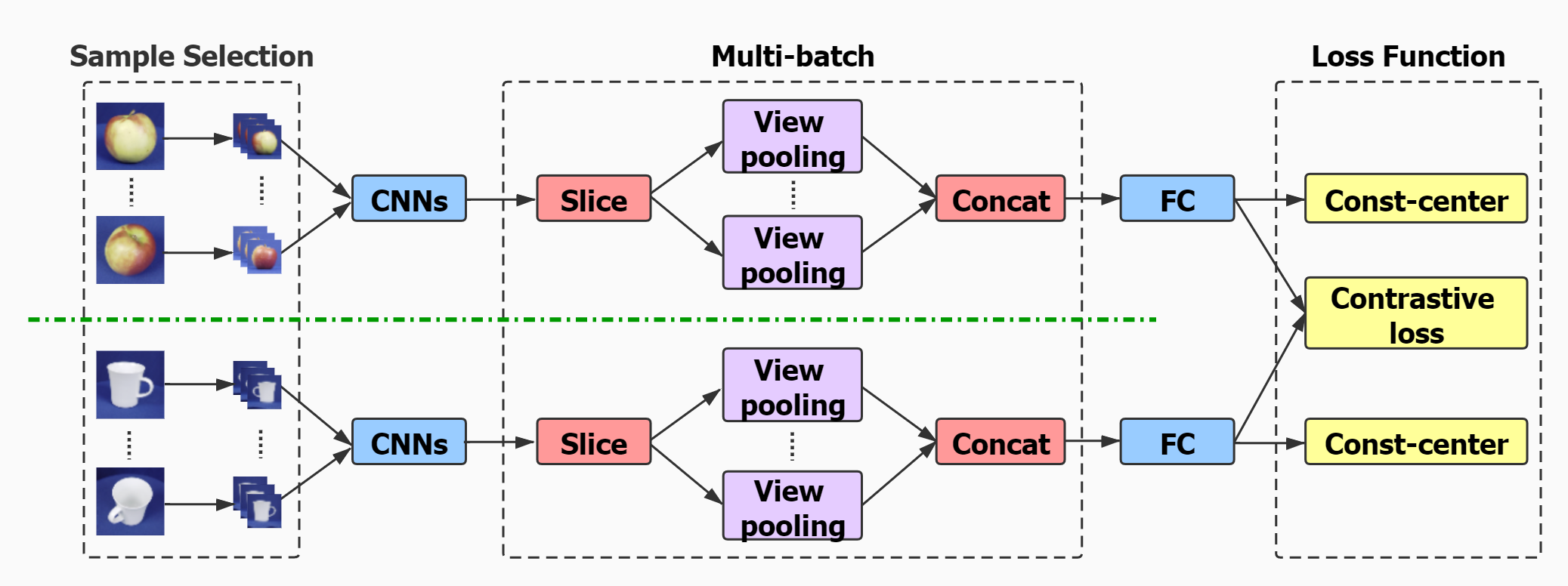}
\caption{The \textbf{MDPCNN} framework for 3D object retrieval. Multiple sample pairs are first chosen by clustering sample selection scheme, and then multiple batch samples are simultaneously inputted in each iteration. Moreover, these samples pass into five convolution layers and enter the Slice layer to separate into feature maps of a single view. A group of feature maps is obtained from the feature maps of every three views through the view pooling layer, and the Concat layer reconstitutes them into the feature maps of multiple batches. Finally, the fully connected layer is used to extract the feature and then three loss functions are calculated separately (best view in color).}
\end{center}
\end{figure*}

\begin{figure}[t]
\begin{center}
\includegraphics[width=3.3in,height = 2.2in]{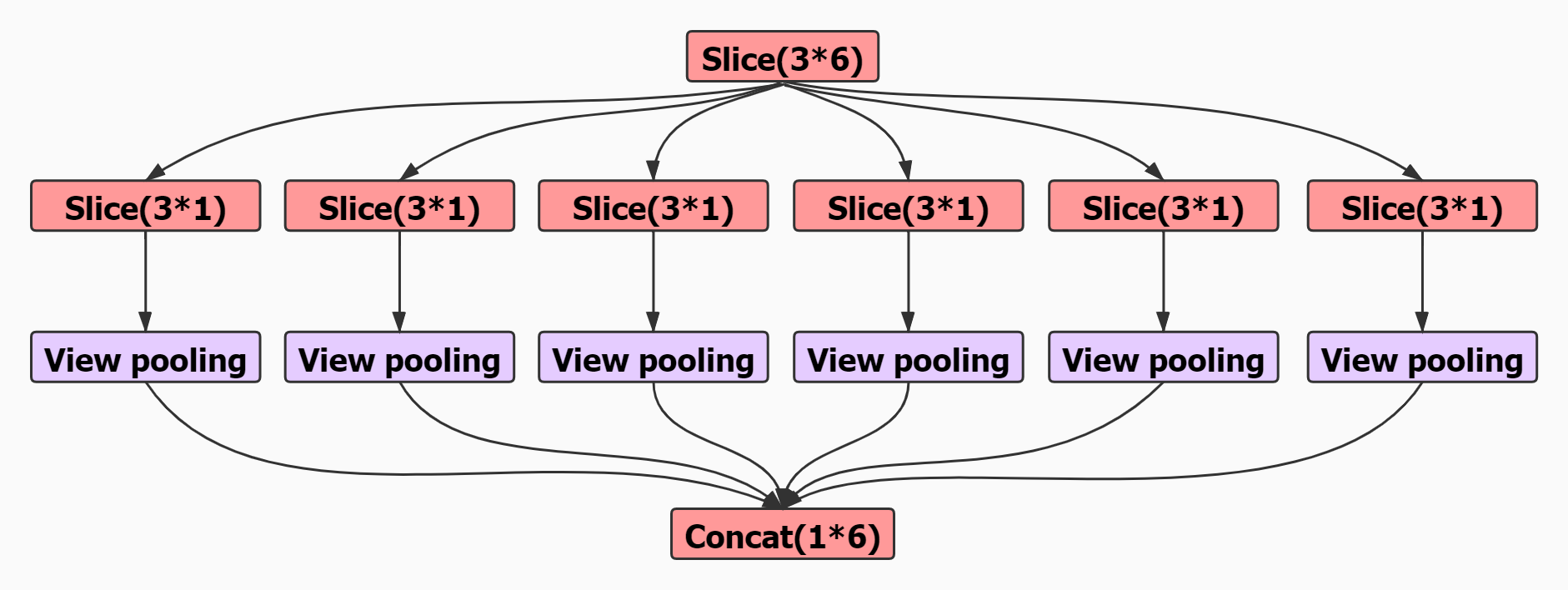}
\caption{A simple example for showing the slice layer how to slice the feature mapping to view pooling layer. Suppose the batch size is 6, and each batch contains one sample pair that contains six views. In other words,  six batches with a total of 18 views are fed to each chain, and each batch contains three multi-view images. Firstly, a Slice layer is used to separate the output feature maps of the upper layer into six batch feature maps, and then six Slice layers are used to separate the feature maps of each batch into a single view feature maps. The feature maps of all views in each batch is calculated through the view pooling layer to calculate the maximum value of the local area elements in the corresponding feature map. Finally,  the feature maps from different view pooling layers are concatenated to form the feature vector (best view in color).}
\end{center}
\end{figure}

\subsection{Clustering Sample Selection}
Choosing suitable training samples is critical to the network training, and even can speed up the convergence of the network. In this work, the deep learning feature for each sample in the dataset is first extracted by VGG-16 network, and then these feature vectors are clustered to get their corresponding class centers. For every object, the next step is to calculate the Euclidean distance between each sample in the object and its corresponding class center to find samples that are farther away from the class center. Thus, these samples with the most difficult to identify are chosen to form a group. In detail, the function can be defined as follows:
\begin{equation}
 D(x_i,center)= argmax_{i=1}^{n_i}\{\|x_i-c_{y_i}\|_2^2\}
 \end{equation}

Where $n_i$ indicates the number of training samples in $y_i$-th category. $x_i\in R^d$ represents the eigenvector of the $i^{th}$ training sample. $y_i$ is the label of $x_i$.  $c_{y_i}$ represents the $j^{th}$ class center. $D(x_i,center)$ denotes the Euclidean distance between $x_i$ in the object and its corresponding class center. In the experiments, these samples with the top $k$ of the farthest samples are chosen. In order to simultaneously mine the latent relationships among views and balance the computation requirements, three samples are selected for each object in the dataset to form the group, thus, we can obtain many groups for the dataset, which have very good representativeness. The merit of the sample selection scheme is that samples with the most difficult to classify for training are the most representative, but it is only a little part of all samples, thus, it can avoid the time and space pressures of training the entire dataset, more importantly, it can improve the convergence speed and enhance the distinctiveness of feature learning.

\subsection{Pairwise Sample Generation}
Since the deep learning network architecture often need to be trained by large scale training samples, but in some real conditions,  amount of training samples is very restricted, thus, the deep learning network architecture cannot be fully trained, thus, the performance of it is very bad. Fortunately, the pairwise sample generation scheme is proposed to solve this issue. Concretely,  we generate positive and negative sample pairs by pairing any two groups in all groups. If these two groups come from the same class, the group pair is a positive sample (label 1), otherwise negative sample (label 0). The pairwise sample generation scheme is shown in Fig.3. It is worth mentioning that when entering the group pair, the label of each group itself and the label of the group pair should be input at the same time.

\begin{figure}[t]
\begin{center}
\includegraphics[width=3.3in,height = 1.9in]{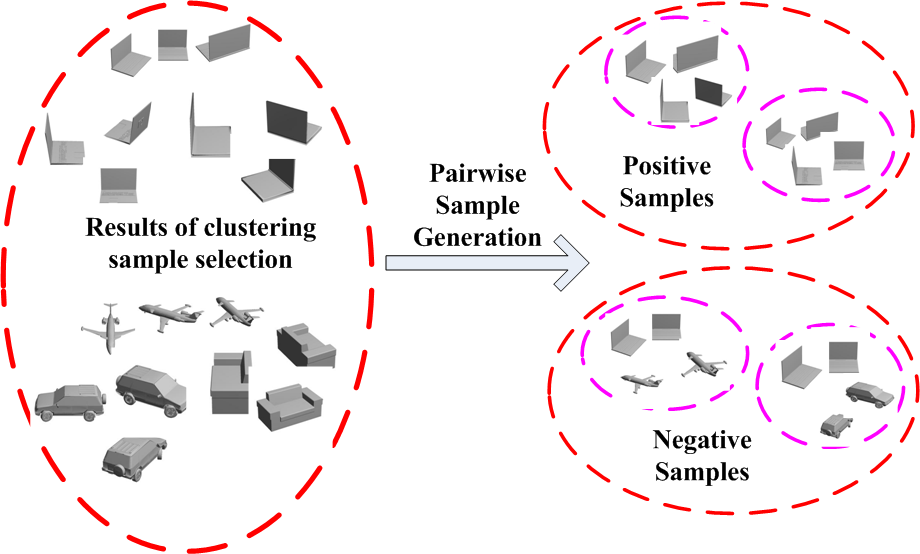}
\caption{Pairwise Sample Generation Scheme. The left ellipse are the results of clustering sample selection, and then we randomly choose two groups from the results to build the pairwise samples.  The right ellipse shows the results of pairwise samples (best view in color)}
\end{center}
\end{figure}

\subsection{Multi-Batch}
If we compare a network to a building, then the layer is the floor, and the blob is the brick. In the network, the data between each layer is transmitted in the form of blobs, including forwarding raw data and reverse gradient information diff. It is a four-dimensional array (Num, Channels, Height, Width), and it can also be written as $(N, K, H, W)$. If we look at Caffe network architecture, the size of $N$ is the same as the size of the batch. $K$ is the same size as the output of each layer. $H$ and $W$ denote the dimensions of the output feature map for each layer. Batch size is an important parameter in machine learning. The increasing batch size can improve memory utilization and parallel efficiency of big matrix multiplication. Within a certain range, in general, the larger the batch size, the more accurate the determined direction of decline is, and the smaller the training oscillation is.

In MVCNN (\cite{Su2016Multi}), multi-view samples are simultaneously inputted the network, but it actually takes up the position of Num (batch size) in the blob. In other words, the batch size of MVCNN is 1, and its size cannot be modified. In order to solve this issue, GPCNN (\cite{Gao2018GroupPair}) was proposed where the external codes were used to set the batch size, but when different datasets are utilized, the external codes need to be modified each time to meet new requirements. Thus, this method is also very inconvenient. In order to further solve this issue, some extra layers, we called Slice layer and Concat Layer, are designed, and are added into the network in the "prototxt" file. The location of them in \textbf{MDPCNN} can be seen in Fig.1. The role of the Slice layer is used to split the input into multiple outputs as needed, and then the feature maps from multiple perspectives in each group are separated into a single view through the Slice layer. Moreover, after obtaining the mapping feature images for each view, the locally optimal screening of multiple views are calculated by the view pooling layer, and then we calculate the maximum value for each element that comes from each mapping image among different views, finally, the max pooling is employed in sub-sampling. In addition, the Concat layer is utilized to splice two or more feature maps on the Channel or the Num dimension. In this way, we can obtain the feature map of multiple batches after being filtered. Since the Slice layer and the Concat layer are not calculated, so we can easily realize multi-batch and multi-view simultaneously without consuming extra time. In addition, when we want to train the model on a new dataset, if the batch size is suitable, thus, we only need to set the path of it. However, if we want to modify the batch size, we only need to easily adjust the number of Slice and Concat layers. More important, by this multi-batch, the convergence of the network can be speeded up, and the convergence accuracy of the network can be improved.

\subsection{Discrimination Loss Function}

The loss function in deep learning is the "baton" of the whole network model, which guides network parameters learning through error back propagation generated by the prediction sample and the real sample marker. If all data is trained to be separable, and only part of the data will cause two different identities to collide with each other because of the fat problem, resulting in false acceptance and false rejection as shown in Fig.4. Our goal is to try to prevent these two situations from appearing. Since the weak labels (only positive or negative label) are utilized, contrastive loss takes restricted relationships into consideration. On the other hand, a contrastive-center loss does not explicitly consider similarity measurements. Therefore, contrastive loss and contrastive-center loss are jointly employed as the optimization objective to make the network more distinguishable, which can enhance the distinctiveness of the features. It can be defined as follows:
 \begin{equation}
  \begin{aligned}
   L_{discrimination} = \alpha\times L_{const}+\beta\times L_{const-center},
  \end{aligned}
\end{equation}
where $\alpha$ and $\beta$ are the control parameters to balance the contribution of the corresponding items. $L_{const}$ and $L_{const-center}$ are two kinds of loss functions. We will introduce them separately.

\begin{figure}[t]
\begin{center}
\includegraphics[width=3.in,height = 2in]{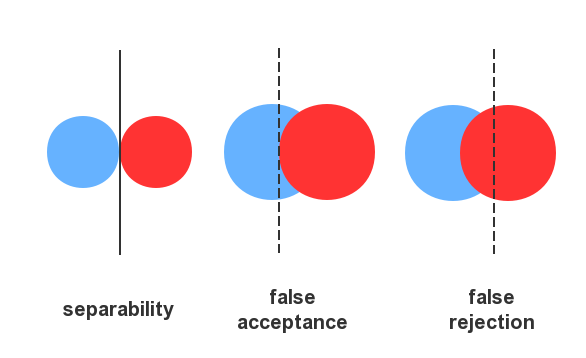}
\caption{Three cases of data separation.(best view in color)}
\end{center}
\end{figure}

\begin{itemize}

\item I) Contrastive Loss, in contrastive loss, two feature vectors are input, and then the distance between them is calculated. In fact, we hope the distance between samples in the same category is as small as possible, but the distance between samples in the different categories is as far as possible. Thus, the expression of the contrastive loss is defined as follows:
 \begin{equation}
 L_{const}= \frac{1}{2N} \sum_{i=1}^N \{y^id^i+(1-y^i)max(m-d^i,0)\},
 \end{equation}
where $d^i=\|a^i-b^i\|^2$ represents the Euclidean distance between two feature vectors $a$ and $b$ that are extracted from the pairwise samples, and $m$ denotes the maximum distance boundary in the current batch. It is used to control the distance between samples from different categories. If the distance is smaller than the maximum distance boundary value, the distance is kept, or it is ignored. $N$ denotes the number of pairwise samples in the current batch. $y^i$ indicates the label of whether the pairwise samples belong to the same category or not. When $y^i$ is set to 1, the Euclidean distance in the feature space of them should be small, but if it is large, thus, the current model is not good, so the loss value should be reduced. On the contrary, when $y^i$ is equal to 0, these two samples are dissimilar, but if the Euclidean distance of them is small, thus, the loss value should become larger.

\item II) Contrastive-center Loss, since the intra-class distance is ignored in the contrastive loss, thus, the contrastive-center loss should be designed where intra-class compactness and inter-class separability should be simultaneously assessed by penalizing the contrastive values between: (1) the distances between training samples from the same category and their corresponding category centers should be as small as possible, and (2) the sum of the distances of training samples to their non-corresponding category centers should be as big as possible. The expression of contrastive-center loss is defined as follows:

 \begin{equation}
 L_{const-center}= \frac{1}{2} \sum_{i=1}^n \frac{\|x_i-c_{y_i}\|^2}{{\sum_{j=1,j\ne y_i}^k\|x_i-c_j\|^2}+\delta},
 \end{equation}
where $n$ denotes the number of training samples in a min-batch. $x_i\in R^d$ represents the eigenvector of the $i^{th}$ training sample. $y_i$ is the label of $x_i$.  $c_{y_i}$  and $c_j$ represents the ${y_i}^{th}$ class center and the $j^{th}$ class center respectively. $k$ denotes the number of class. $\delta$ is a constant to prevent the denominator equal to 0. During each iteration, the distances between training samples to their corresponding category centers are decreased and the sum of the distances between training samples to their non-corresponding category centers is increased by the contrastive-center loss.

\end{itemize}

\subsection{Comparison Against Related Algorithms}

1) \textbf{MDPCNN vs MVCNN (\cite{Su2016Multi})}, although our idea is inspired by the MVCNN, \textbf{MDPCNN} is very different from MVCNN. Firstly, there is single network architecture in MVCNN which has multi-view input, but in \textbf{MDPCNN}, the pairwise network architecture is employed and each network has multi-view input. Secondly, in MVCNN, the large number of training samples are needed, and its performance will be bad on small-scale size dataset. However, since the pairwise CNN that is similar to Siamese Convolutional Neural Network, is employed in \textbf{MDPCNN}, thus, even if the amount of training samples is very restricted in the original dataset, but the plenty of pairwise samples can be generated, which can be used to fully train \textbf{MDPCNN}. Thirdly, in MVCNN, the Softmaxloss function is employed, but in \textbf{MDPCNN}, the contrastive loss and contrastive-center loss are jointly used. Fourthly, in \textbf{MDPCNN}, the batch size can be easily modified than that of MVCNN which also can speed up the convergence of the network and improve its convergence accuracy. Finally, the sample selection scheme is proposed in \textbf{MDPCNN}, by this way, these samples that are more difficult to identify are chosen, and it is beneficial to both convergence rate and final performance, but in MVCNN, there is no sample selection. Thus, \textbf{MDPCNN} is obviously different from MVCNN.

2) \textbf{MDPCNN vs Siamese Convolutional Neural Network (\cite{Chopra2005Learning,Norouzi2012Hamming})}, the pairwise network is employed in both \textbf{MDPCNN} and Siamese Convolutional Neural Network, but these two network architectures are very different. Firstly, in \textbf{MDPCNN}, three or more images can be entered, but in Siamese Convolutional Neural Network, only an image can be input into one network at a time; Secondly, different pooling schemes are utilized in both networks. In the former, the view pooling scheme is employed, which is utilized to mine the latent relationships among multiple views, but in the latter, max pooling or average pooling schemes are often used. Thirdly, the loss function is also very different. In the former, the contrastive loss and contrastive-center loss are jointly used, but in the latter, the Euclidean distance function is employed. Thus, these two networks are obviously various.

3) \textbf{MDPCNN vs GVCNN (\cite{Feng2018GVCNN})}, In GVCNN, the intrinsic hierarchical correlation and discriminability
among views are well exploited, which also belongs to a single network, but there is multi-view input for it. However, in \textbf{MDPCNN}, the pairwise network architecture is employed, and for each network, multi-view images are also entered. In addition, the loss function is also very different, and the sigmoid function is used in GVCNN, but in \textbf{MDPCNN}, the contrastive loss and contrastive-center loss are jointly employed which is very helpful for improving the discriminability of extracting feature.

In total, our \textbf{MDPCNN} are very different from the existing works.

\section{Experiments}
\label{sec:experiments}

In order to assess the retrieval performance of \textbf{MDPCNN}, four public 3D object datasets are used to perform 3D object retrieval experiments. In the following, the datasets and evaluation criteria are first given, and then, the experimental setting is introduced. Meanwhile, we will evaluate \textbf{MDPCNN} from three aspects respectively: (1) We will compare the performance of \textbf{MDPCNN} on different datasets with existing methods. (2) We will analyze the importance of each part in \textbf{MDPCNN}. (3) We will show its loss convergence with different batch sizes and convergence speed in \textbf{MDPCNN}.

\subsection{Dataset}
In our experiments, four widely-used datasets are employed where each object (or each model) in these datasets is projected by a free set of virtual cameras, and then these views are employed to represent the object. The details of these datasets are introduced as follows:
\begin{itemize}
 \item ETH 3D object dataset (\cite{ess2008a}), in this dataset, there are 80 objects that belong to 8 categories, and 41 different view images are used to represent each object from ETH dataset. In total, it contains 3280 views belonging to 8 categories.
\item NTU60 3D model dataset (\cite{Chen2003On}), in this dataset, there are 549 objects that belong to 47 categories, and
60 different view samples are utilized to describe each object from  NTU60 dataset.  In total, it contains 32940 views belonging to 47 categories.
 \item MVRED 3D category dataset (\cite{Liu2016Multi}), in this dataset, there are 505 objects that belongs to 61 categories, and 36 different view images are included in each object from MVRED dataset. In total, it contains 18180 views belonging to 61 categories.
 \item Modelnet40 3D category dataset (\cite{Su2016Multi,Wu20153D}). ModelNet (\cite{Wu20153D}) is very popular 3D model dataset, which is released on the Princeton ModelNet website.\footnote{The Princeton ModelNet. Http://modelnet.cs.princeton.edu/}. As for Modelnet40, it is a 40-category well-annotated subset of Modelnet, which also can be downloaded in the same website. In all experiments, we strictly follow the training and test split of ModelNet40 as in \cite{Wu20153D,Su2016Multi}. In the training set, the amount of objects in each category changes from 64 to 80, but in the test set, the amount of objects in each category is 20. In both datasets, the number of views in each object is fixed to 12. In total, it contains 38196 views belonging to 40 categories.

\end{itemize}

some examples from ETH, NTU60, MVRED and Modelnet40 datasets are given in Fig.5.

\begin{figure}[h]
\begin{minipage}{0.49\linewidth}
  \centerline{\includegraphics[width=1.6in,height = 1.6in]{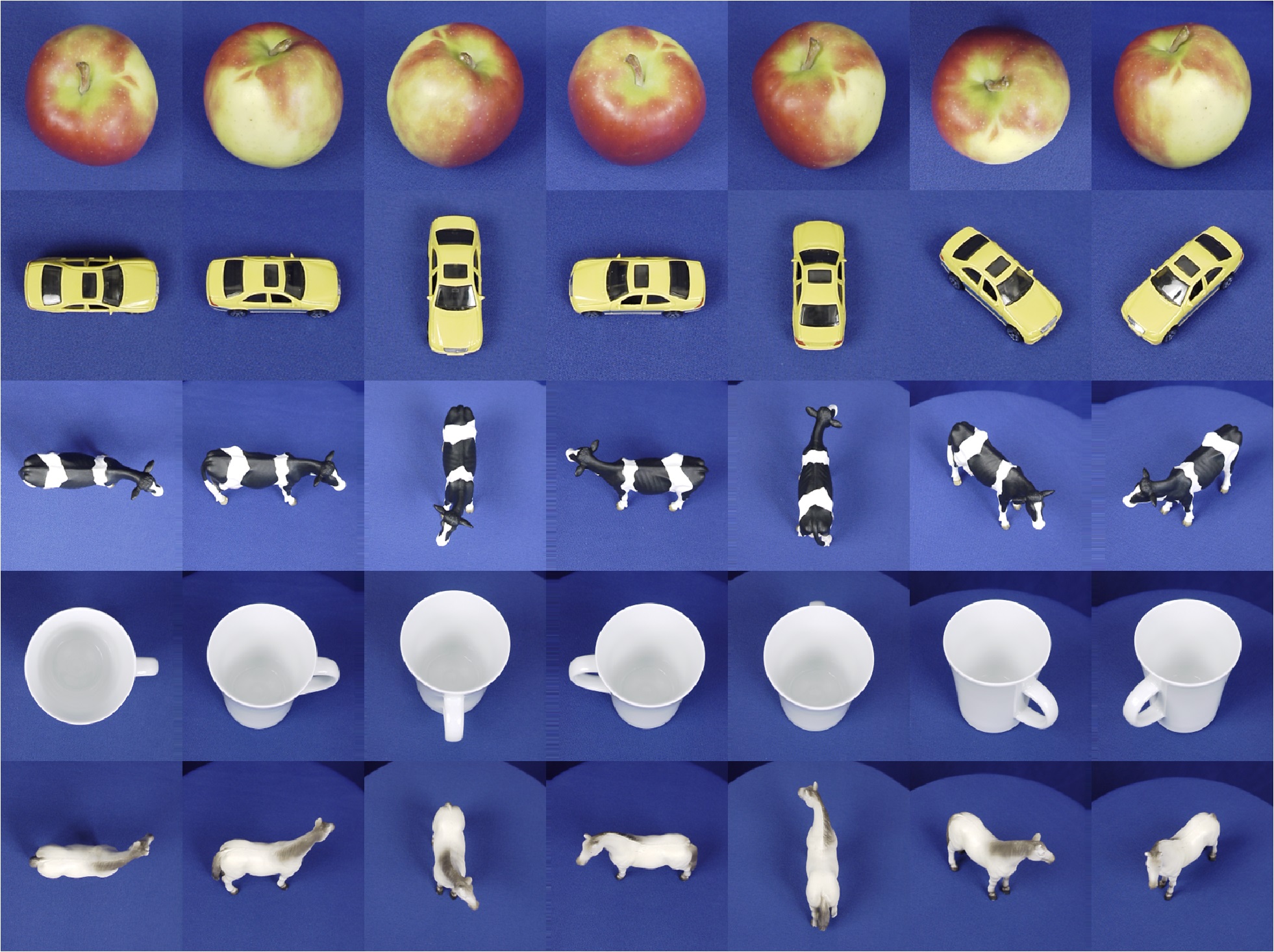}}
\end{minipage}
\hfill
\begin{minipage}{0.49\linewidth}
  \centerline{\includegraphics[width=1.6in,height = 1.6in]{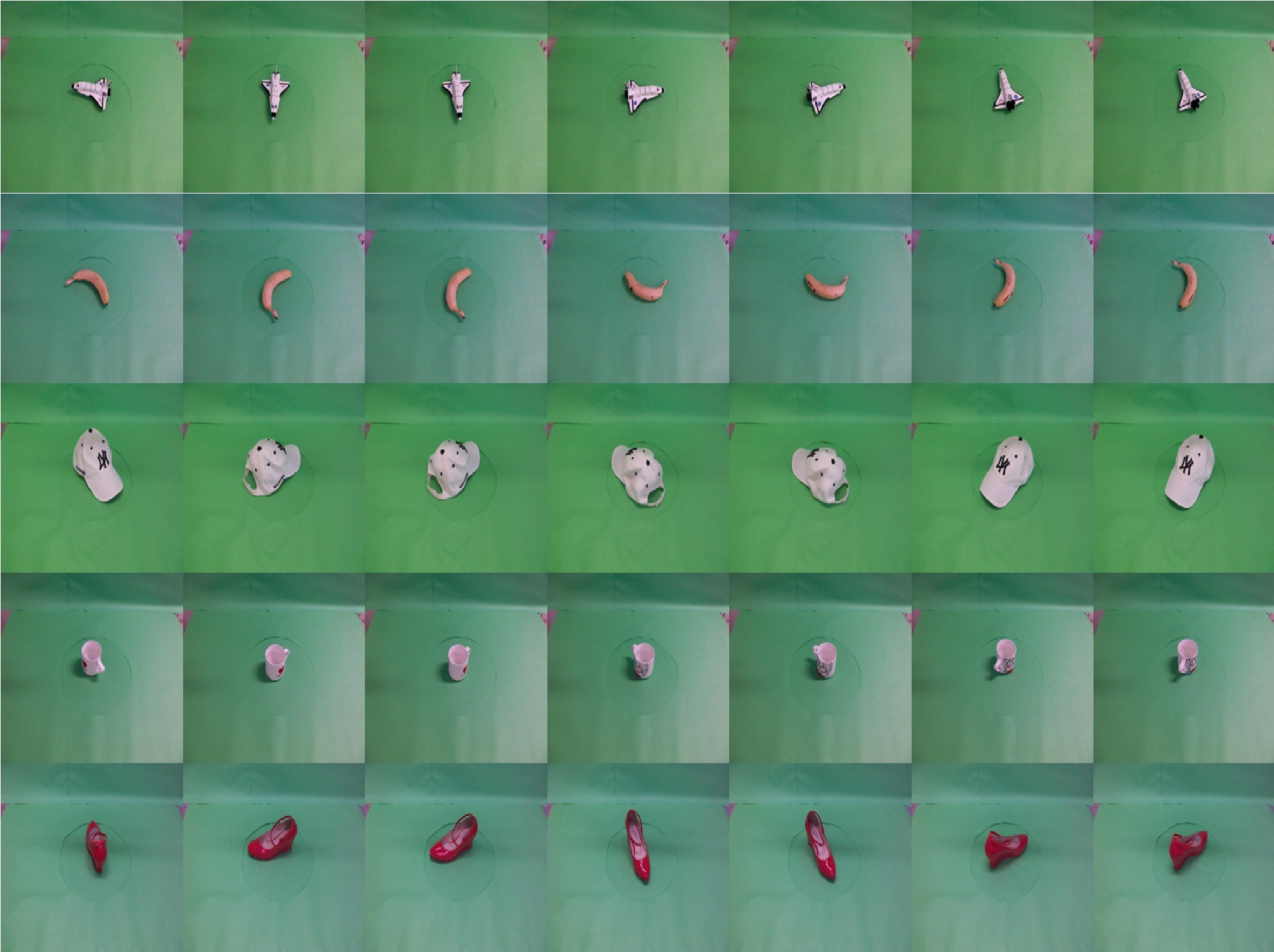}}
\end{minipage}\\
\vfill
\begin{minipage}{0.49\linewidth}
  \centerline{\includegraphics[width=1.6in,height = 1.6in]{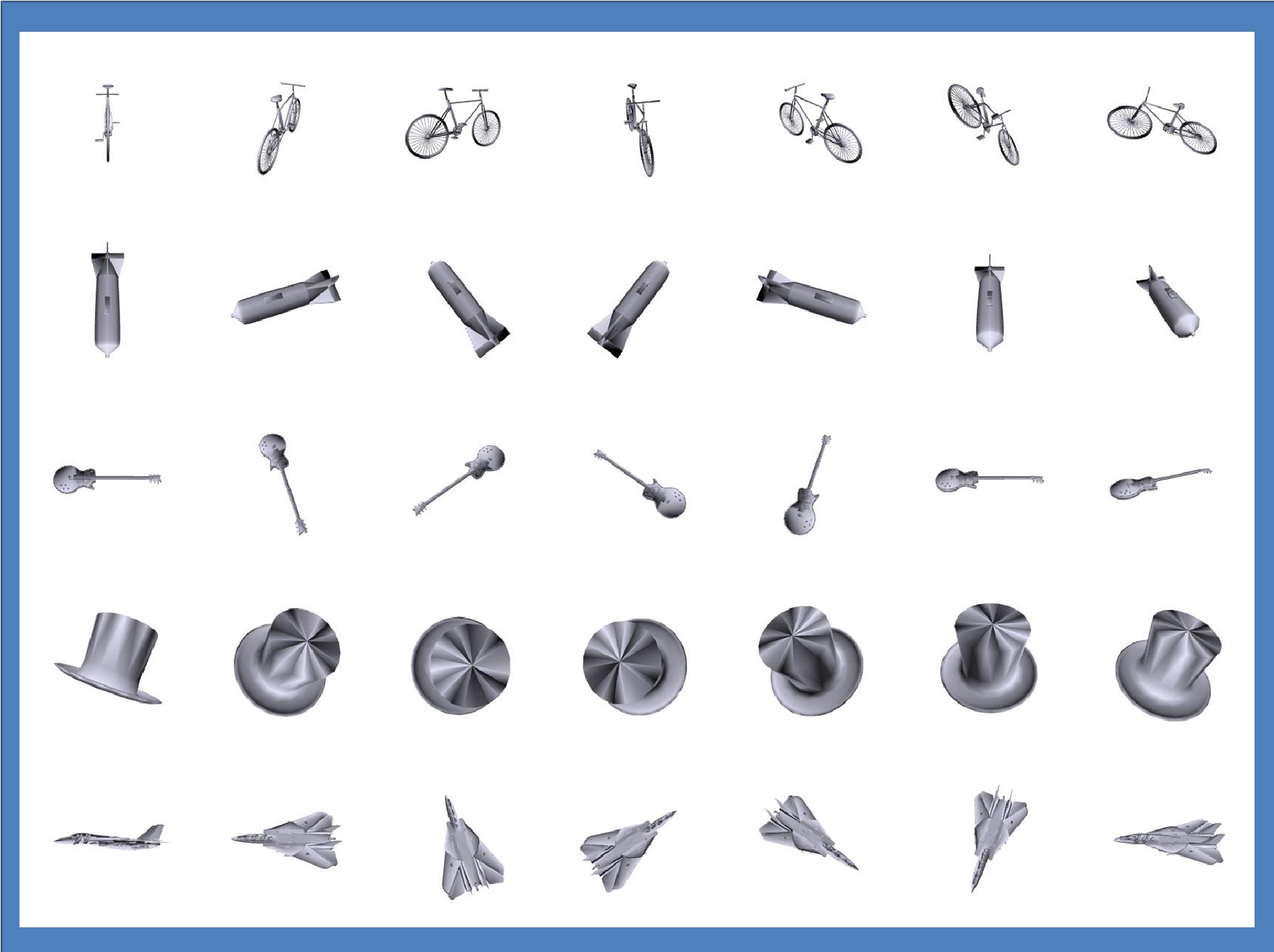}}
\end{minipage}
\begin{minipage}{0.49\linewidth}
  \centerline{\includegraphics[width=1.6in,height = 1.6in]{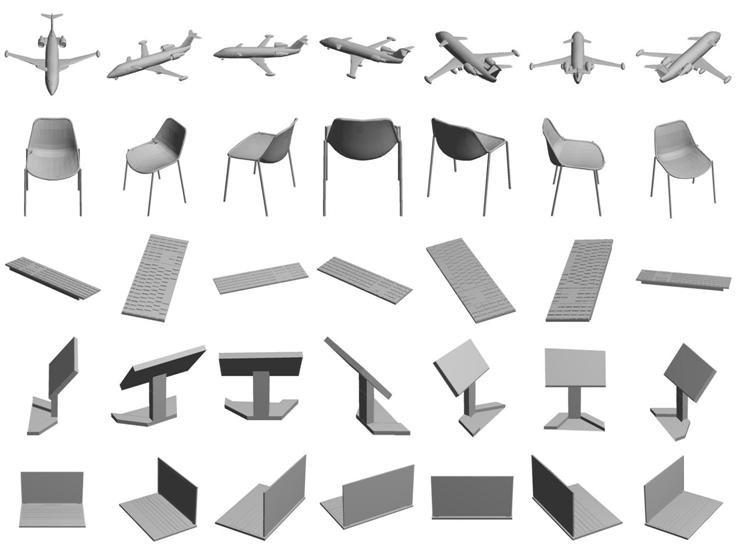}}
\end{minipage}
\caption{ Images from different views and different datasets are shown. From top to down, and left to right, samples of each block are chosen from ETH, NTU60, MVRED and ModelNet40 datasets, respectively}
\end{figure}

\subsection{Evaluation Criteria}

Since different evaluation criterions are employed in different datasets, thus, for a fair comparison, different evaluation criterions are also utilized according to related works (\cite{Liu2016Multi,Chen2003On,ess2008a,Su2016Multi}). Concretely, in EHT, NTU60 and MVRED datasets, seven evaluation criterions are utilized to assess its performance, and they are Nearest Neighbor (NN), First Tier (FT), Second Tier (ST), F-measure (F), Discounted Cumulative Gain (DCG) (\cite{osada2001matching}), Average Normalized Modified Retrieval Rank (ANMRR) and Precision-Recall Curve (PR-Curve). For  details, you could find them in (\cite{Liu2016Multi,Chen2003On,ess2008a}). In ModelNet40 dataset, mean average precision (mAP) (\cite{Su2016Multi}) is used.

Here, it is noted that for NN, FT, ST, F, DCG and mAP.  The bigger the better, but for ANMRR the smaller the better, and the greater of the area under PR-Curve, its performance is better. In addition, in order to fair competition, we also follow the parameter setting in other papers, where $k$ is set to ten, thus, we also compute the performance when $k$ is used (each retrieval, top ten retrieval results are utilized to compute the evaluation criteria value).

\subsection{Competing Methods}
Several popular algorithms are implemented for comparisons:
\begin{itemize}

 \item Adaptive View Clustering (AVC) (\cite{osada2001matching}).  The optimal 2D characteristic views of a 3D model based on the adaptive clustering scheme is selected in AVC, and then a probabilistic Bayesian scheme is utilized for 3D model retrieval.

\item Camera Constraint Free View (CCFV) (\cite{Gao2012Camera}). The positive matching model and the negative matching model are first built, and then the query Gaussian models are generated by combining them, finally, on the basis of it,  CCFV model is generated.

\item Weighted Bipartite Graph Matching (WBGM) (\cite{Gao20113D}). WBGM builds the weighted bipartite graph only with the attributes of individual 2D views.

\item Hausdorff distance (HAUS) $\&$ Nearest Neighbor (NN) (\cite{Steinbach2000A}). The distances between a set and its nearest point in the other set are first calculated, and then the Hausdorff distance is employed to find the maximum distance of them. The nearest neighbor-based method is used to obtain the minimum distance of them.


\item Siamese Convolutional Neural Network (\cite{Chopra2005Learning,Norouzi2012Hamming}). Siamese Convolutional Neural Network takes a pair of samples instead of taking single sample as input, and the loss functions are usually defined over pairs.

\item Group-Pair Convolutional Neural Network (GPCNN) (\cite{Gao2018GroupPair}). GPCNN constructs an end-to-end double-chain network with multi-view input, and guarantees the number of training of small datasets by group-pair approach.

\item Multi-view CNN (MVCNN) (\cite{Su2016Multi}). In MVCNN, there is single network architecture in MVCNN, but multiple views of the shapes are provided into the network, and then view pooling scheme is utilized to mine the relationships among these views.

\item Group-view CNN (GVCNN) (\cite{Feng2018GVCNN}). In GPCNN, a hierarchical view-group-shape architecture is proposed, and then all group level descriptors are weighted embedded to generate the shape level descriptor.

\item CNN+LSTM (\cite{Ma2019Learning}). In CNN+LSTM, the convolutional neural networks (CNNs) are combined with long short-term memory (LSTM) to exploit the correlative information from multiple views.

\end{itemize}

\subsection{Experimental Settings}

For ETH, NTU60 and MVRED dataset, the first 80\% views of each object are used as training dataset and other views are considered as testing dataset. As for Modelnet40, the training and test split of ModelNet40 in (\cite{Wu20153D,Su2016Multi}) is employed. Since the amount of samples in training and testing process is very limited, thus, clustering sample selection scheme is firstly used to choose these samples of the most difficult to identify, and then the pairwise sample generation scheme is utilized to obtain pairwise samples. Concretely, in order to balance performance and memory requirements, 10k positive pairwise samples and 30k negative pairwise samples are built in ETH dataset during the training phase. Meanwhile, 30k positive pairwise samples and 90k negative pairwise samples are generated on MVRED, NTU60 and ModelNet40, respectively. Since the number of objects in ModelNet40 is much bigger than that of ETH, MVRED and NTU60 datasets, thus, in order to fully consider much more situations, 90k positive pairwise samples and 180k negative pairwise samples are generated for ModelNet40 dataset. At the testing stage, we randomly select the remaining three views from each object to form a group which is utilized to represent the object, and then these views are used to retrieve objects from the gallery dataset. We repeat ten times experiments, and then the average result is used as the final performance. In fact, we have tried to select much more views from each object, such as 5 views or 7 views, but we found that the memory requirements were too large, and MDPCNN could not be fully trained. However, we also want to mine the latent relationships among different views, thus, 3 views can balance the latent relationships and the memory requirements.

\textbf{1) Training}. Our network consists of five convolution layers, n (batch size) Slice layers and view pooling layers, a Concat layer, two fully connected layers and three loss layers. The sizes of convolution kernels of each convolution layer are 32, 64, 128, 256 and 512, respectively in MVRED, NTU60 and ModelNet40 datasets, while ETH is 16, 32, 64, 128 and 256, respectively. The batch size is set to 12 for ETH, 40 for MVRED, NTU60 and ModelNet40. The mini-batch stochastic gradient descent (SGD) is employed to update the network parameters. The learning rates of MVRED, NTU60 and ModelNet40 are initialized as 0.05 (ETH is 0.01), and every third epoch is reduced by 0.5 times.  The weight of the contrastive loss is set to 0.99 ($\alpha = 0.99$), and the weight of two contrastive-center losses is set to 0.01 ($\beta = 0.01$) in all four datasets. Besides, $m$ is set to 1 in the contrastive loss,  and $\delta$ is set to 1. Finally, all the gradients that are produced by every objective separately are computed, and then the weighted gradients are added together to update the network.

\textbf{2) Testing}. The group-pair in the test phase (query samples) is composed of the group of each object and the group of other objects in the gallery set. The last fully connected layers of the two chains are utilized to extract the features as the descriptors of these two objects. Next, the Euclidean distance between them is calculated. Finally, the retrieval results are sorted, and different evaluation criteria are computed.

\subsection{Large Scale Pairwise Samples Generation}

The numbers of objects and views in these 3D datasets are given in Table 1. From Table 1, we can find that the numbers of objects in ETH, NTU60, MVRED and ModelNet40 datasets are 80, 549, 505 and 3183, respectively, thus, it will be difficult for fully training deep Convolutional Neural Network. In order to generate large scale samples to fully train CNN models, pairwise sample generation scheme is proposed in Fig.3. By this way, the numbers of pairwise samples in ETH, NTU60, MVRED and ModelNet40 datasets can reach 33,685,600, 2,601,768,096, 416,903,760 and 1,114,113,660, respectively, whose increments have reached more than 421,070-fold, 4,739,104-fold, 825,552-fold and 350,020-fold respectively. In consideration of computer memory, 40,000, 120,000, 120,000 and 120,000 pairwise samples are only generated for ETH, MVRED, NTU60 and ModelNet40 datasets, respectively. Since there are plenty of training samples, thus, \textbf{MDPCNN} can be effectively trained.

In order to assess its performance, we will evaluate it on four different 3D object datasets, but since different evaluation criterions are employed on different datasets. In fact, on ETH, NTU60 and MVRED datasets, NN, FT, ST, F, DCG and ANMRR are used (\cite{Liu2018View,Gao2018GroupPair,Gao2016A,Gao2012Camera,Gao20113D}), but on ModelNet40 dataset (\cite{Feng2018GVCNN,Su2016Multi,Wu20153D,Zhaoxiang2016GIFT}), mAP is utilized.  Thus, the performance evaluation of \textbf{MDPCNN} on ETH, NTU60 and MVRED datasets are put together, but the performance evaluation of \textbf{MDPCNN} on ModelNet40 is separately discussed.

\begin{table*}
\caption{Large scale samples generated by the pairwise sample scheme on different datasets}
\begin{center}
\begin{tabular}{cccccc}
  \hline
           \multirow{2}{*}{Datasets} & \multicolumn{3}{c} {Training Dataset }& \multicolumn{2}{c} {Pairwise Samples Generation} \\
\cline{2-6}
      &Objects &Views per Object &Samples in total &The number of Views &Pairwise Samples in total\\
\hline
ETH & $80$ &  $32$ & $2560$ & $3$  & $C_{32}^3 \times C_{80}^2 = 33,685,600 $\\
\hline
NTU60 & $549$ &  $48$ & $26352 $& $3$  &$C_{48}^3 \times C_{549}^2 = 2,601,768,096 $\\
\hline
MVRED & $505$ &  $28$ & $14140$ & $3$ & $C_{28}^3 \times C_{505}^2 = 416,903,760 $\\
\hline
ModelNet40 & $3183$ &  $12$ & $38196$ & $3$ & $ C_{12}^3 \times C_{3183}^2 = 1,114,113,660 $\\
\hline
\end{tabular}
\end{center}
\end{table*}

\subsection{Performance Evaluation on ETH, MVRED and NTU60 Datasets}
In this section, we will assess the performance of \textbf{MDPCNN} on three datasets, and then the state-of-the-art methods also are evaluated. Meanwhile, these methods will be compared to these datasets with different evaluation criterions. Since all views of one object are input into the network, but we can observe from Table 1 that the number of objects in ETH, NTU60 and MVRED datasets is very limited, and the small-scale sample problem need to be solved. Large-scale experiments demonstrate that the deep learning features outperform hand-crafted features, but large scale training samples are often required in deep learning methods. Thus, in our experiments, MVCNN and GPCNN are not assessed on these three datasets. In AVC, CCFV, WBGM, HAUS and NN, the default settings are also used, and Zernike feature is utilized to describe each sample. In Siamese network and GPCNN, the default settings are employed, and ImageNet1K dataset is utilized to pre-train the network, and then they are fine-tuned, respectively. The experimental results on ETH, MVRED and NTU60 datasets are given in Fig.6, Fig.7 and Fig.8, respectively. From them, we can observe that:

1) In these traditional 3D object retrieval algorithms, the feature extraction and object retrieval are performed, respectively, but the \textbf{MDPCNN} is an end-to-end network that feature extraction and object retrieval are put together. Meanwhile, although the number of training samples is small in the original dataset, amount of training samples are generated by \textbf{MDPCNN}, thus, the requirement of the amount of training samples in the original dataset is small, which can efficiently solve the small-scale sample problem. Thus, in all three datasets, \textbf{MDPCNN} obviously has better performance than that of traditional 3D object retrieval algorithms. For example, in Fig.6 (b),  Fig.7 (b) and  Fig.8 (b), the areas under the curve of \textbf{MDPCNN} are much bigger than that of traditional 3D object retrieval algorithms. In addition, we also find that the performance of Siamese network is not so good. The reason is that although the pairwise network architecture is used that a lot of pairwise samples can be obtained,  multi-view information is missing. Similarly, in GPCNN, the pairwise network and multi-view information are simultaneously employed, thus, its performance on these three datasets are much better than that of these traditional 3D object retrieval methods.

2)  When compared \textbf{MDPCNN} with the Siamese network, in all these three datasets, \textbf{MDPCNN} obviously outperforms the Siamese network, for example, in ETH dataset, the Siamese network is a little better than that of traditional 3D object retrieval methods, but \textbf{MDPCNN} obviously outperforms traditional 3D object retrieval methods. In the  MVRED and NTU60 datasets, the performances of  traditional 3D object retrieval methods are better than that of the Siamese network, but \textbf{MDPCNN} still obviously outperforms traditional 3D object retrieval methods. The key important reason is that multi-view information is ignored in the Siamese network.

 3) When compared \textbf{MDPCNN} with GPCNN in ETH dataset, it has certain advantages in the first four sets of indicators, but the performance on the latter two sets of indicators is slightly worse. This may be because GPCNN has achieved good performance by randomly selecting samples, but this may cause unstable network performance, but it is easy to see from the PR curve in Fig.6 that \textbf{MDPCNN} is still better than that of GPCNN overall. Fig.6 shows \textbf{MDPCNN} can achieve a gain of 2\%-32\%, 4\%-42\%, 10\%-29\%, and 1\%-13\% on NN, FT, ST and F, respectively.

 4) Similarly, when we compared \textbf{MDPCNN} with GPCNN in MVRED and NTU60 datasets, the performance improvement is also very obviously. On the whole, Fig.7 shows \textbf{MDPCNN} outperforms 25\%-69\%, 34\%-65\%, 21\%-64\%, 21\%-51\% and 33\%-72\% on NN, FT, ST, F and DCG, respectively and get a decline of 28\%-66\% on ANMRR. In Fig.8, \textbf{MDPCNN} outperforms 23\%-65\%, 30\%-56\%, 12\%-54\%, 25\%-51\% and 32\%-64\% on NN, FT, ST, F, DCG, respectively and get a decline of 24\%-57\% on ANMRR.

\begin{figure}[h]
\begin{minipage}{1\linewidth}
  \centerline{\includegraphics[width=3.2in,height = 2in]{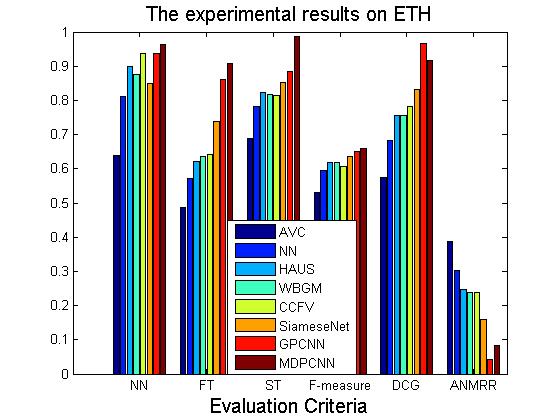}}
  \centerline{(a) Retrieval Comparison }
\end{minipage}
\hfill
\begin{minipage}{1\linewidth}
  \centerline{\includegraphics[width=3.2in,height = 2in]{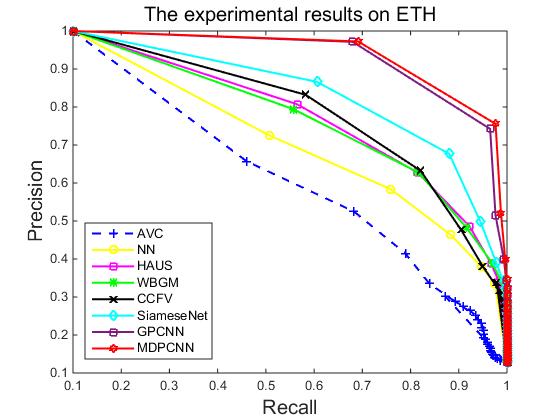}}
  \centerline{(b) PR Comparison}
\end{minipage}
\caption{Retrieval result comparison and PR Curve comparisons with state-of-the-art methods on ETH dataset}
\end{figure}

\begin{figure}[h]
\begin{minipage}{1\linewidth}
  \centerline{\includegraphics[width=3.2in,height = 2in]{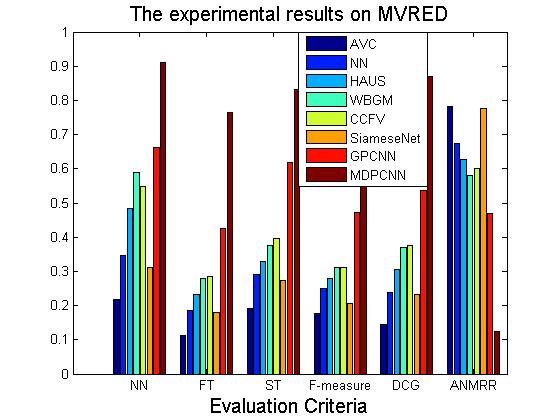}}
  \centerline{(a) Retrieval Comparison }
\end{minipage}
\hfill
\begin{minipage}{1\linewidth}
  \centerline{\includegraphics[width=3.2in,height = 2in]{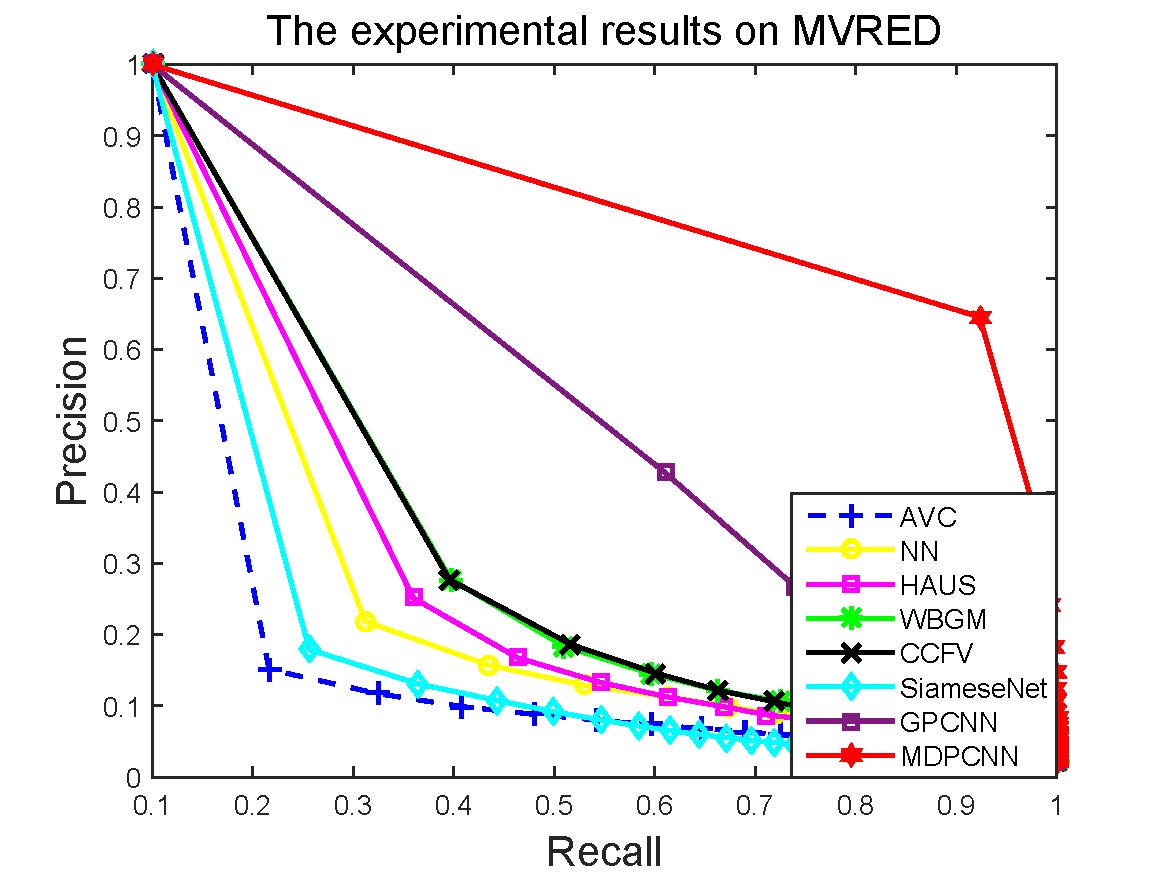}}
  \centerline{(b) PR Comparison}
\end{minipage}
\caption{Retrieval result comparison and PR Curve comparisons with state-of-the-art methods on MVRED dataset}
\end{figure}

\begin{figure}[h]
\begin{minipage}{1\linewidth}
  \centerline{\includegraphics[width=3.2in,height = 2in]{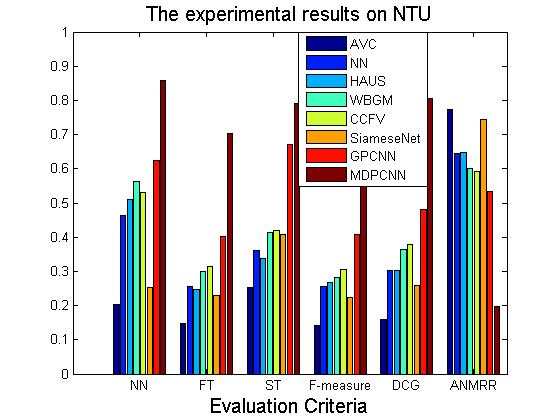}}
  \centerline{(a) Retrieval Comparison }
\end{minipage}
\hfill
\begin{minipage}{1\linewidth}
  \centerline{\includegraphics[width=3.2in,height = 2in]{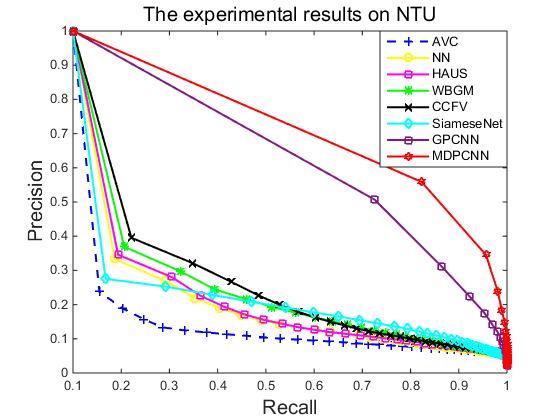}}
  \centerline{(b) PR Comparison}
\end{minipage}
\caption{Retrieval result comparison and PR Curve comparisons with state-of-the-art methods on NTU60 dataset}
\end{figure}

\subsection{Performance Evaluation on ModelNet40 Dataset}

In order to further assess the performance of our \textbf{MDPCNN} method, 3D shape retrieval experiments on the Princeton ModelNet dataset are further conducted, where 127,915 3D CAD models from 622 object categories are included in ModelNet. ModelNet40 is subsampled as the subset of ModelNet, which contains 40 popular object categories. The training/testing split setting in (\cite{Wu20153D,Su2016Multi}) is employed. In our experiments, some popular descriptors that are often used in 3D shape retrieval, are first assessed. And then different CNN networks with different training datasets are also appraised. In our experiments, \textbf{MDPCNN} is compared with 3D ShapeNets (\cite{Wu20153D}),  MVCNN (\cite{Su2016Multi}), GIFT (\cite{Zhaoxiang2016GIFT}),  PANORAMA-NN (\cite{N20220:2017}), GVCNN (\cite{Feng2018GVCNN}), and CNN+LSTM (\cite{Ma2019Learning}). The default parameters setting of these methods is used. The experimental results on ModelNet40 dataset are given in Table 2. From it, we can observe that:

1) The performances of Spherical Harmonics descriptor (SPH), LightField descriptor (LFD), Fisher vectors (FV 12x) and 3D ShapeNets are 33.3\%, 40.9\% 43.9\% and 49.2\%, respectively, which are considered as our baseline. Since they belong to hand-crafted features, feature extraction and retrieval are conducted respectively, thus, their performances are not very good. When CNN network is utilized, the performance of the deep learning feature can obtain some improvement, for example, the performances of CNN (f.t.) and CNN (f.t. 12x) are 61.7\% and 62.8\% respectively, which are much better than that of SPH, LFD, FV and 3D ShapeNets.

2) In CNN, multi-view information is not explored, but in MVCNN, multiple view samples are fully discussed, thus, MVCNN (f.t. 12x) outperforms the CNN (f.t. or f.t. 12x) by a significant margin, for example, the accuracies of MVCNN (f.t. 12x) and MVCNN (f.t, 80x) reach 70.1\% and 70.1\%, respectively whose increment achieves about 8\%. Meanwhile, when the low-rank Mahalanobis metric learning is further studied, the mAP of MVCNN can be further improved. For example, the performances of both MVCNN (f.t.+ metric, 12x) and MVCNN (f.t.+ metric, 80x) attain 80.2\%, whose improvement reaches 10\% to MVCNN (f.t. 12x) and MVCNN (f.t, 80x). Thus, metric learning is also very helpful. In addition, we also find that when 3D model is represented by much more multi-view samples, such as 80, the performance of MVCNN cannot obtain any improvement.

3) Although multi-view information has been explored in MVCNN, the discriminability and the intrinsic hierarchical correlation among views are not discussed, thus, a hierarchical view-group-shape architecture is proposed in GVCNN where all group level descriptors are weighted assembled to generate the shape level descriptor.  Table 2 demonstrates that the performances of GVCNN (f.t. + 8x) and GVCNN (f.t. + metric, 8x) reach 79.7\% and 84.5\%, respectively, but the accuracies of MVCNN (f.t. + 12x) and MVCNN (f.t. + metric + 12x) are 70.1\% and 80.2\%, whose improvements achieve 9.6\% and 4.3\%, respectively. When 12 views are utilized to represent the 3D model, the mAP of GVCNN can further obtain improvement. Thus, the discriminability among views is very useful for 3D object retrieval.

4) In GVCNN, a hierarchical view-group-shape architecture is designed, and then the view-pooling scheme is utilized to generate a group level description. Moreover, the shape level descriptor is generated by combining all group level descriptors according to their discriminative weights. However, in \textbf{MDPCNN}, the discriminability among views are explored by two ways: 1) the clustering scheme is utilized to select the most difficult samples; 2) the discrimination loss function is designed to improve the discriminability of feature representation, and further improve the performance of the network precision. Table 2 shows that the performance of \textbf{MDPCNN (all)} outperforms the GVCNN by 2\%, but when comparing with GIFT (\cite{Zhaoxiang2016GIFT}) (81.9\%) and PANORAMA-NN (83.5\%) (\cite{N20220:2017}), whose improvements reach 5.5\% and 4.1\%, respectively. Meanwhile, when we compare \textbf{MDPCNN} with the baseline, its improvement is very obvious. Besides, we also compare \textbf{MDPCNN} with the CNN+LSTM where 
the LSTM is utilized to exploit the correlative information from multiple views, but we find that the improvement of \textbf{MDPCNN} also can reach 3.3\%.  In a word, \textbf{MDPCNN} outperform state-of-the-art algorithms.

\begin{table*}
\centering
\fontsize{9}{9}\selectfont
\caption{Performance Comparison on the ModelNet-40 Dataset. '\#x' means the number of views in each object}
\begin{center}
\begin{tabular}{lccccc}
  \hline
           \multirow{2}{*}{Methods} & \multicolumn{3}{c} {Training Config.} & \multicolumn{1}{c} {Test Config.} & \multirow{2}{*} {Accuracy (\%)} \\
\cline{2-5}
      &Pre-train &Fine-tune &\#Views  &\#Views \\
\hline
(1) SPH (\cite{Kobbelt2003Rotation}) & -  & -  & - &- & 33.3 \\
(2) LFD (\cite{Chen2010On})  & -  & -  &- &- & 40.9 \\
(3) 3D ShapeNets (CVPR 2015) (\cite{Wu20153D}) & ModelNet40  & ModelNet40 &- &- & 49.2 \\
\hline
(4) FV & -  &ModelNet40  & 12 & 1 & 37.5 \\
(5) FV 12x  & -  &ModelNet40  & 12 & 11 & 43.9 \\
(6) CNN  &ImageNet1K  &-  & - & 1 & 44.1 \\
(7) CNN, f.t  &ImageNet1K  &ModelNet40  &12 & 1 & 61.7 \\
(8) CNN, 12x  &ImageNet1K  &-  &- & 12 & 49.6 \\
(9) CNN, f.t, 12  &ImageNet1K  &ModelNet40  &12 & 12 &62.8 \\
\hline

(10) MVCNN (ICCV 2015) 12x (\cite{Su2016Multi}) & ImageNet1K  &-  &- & 12 & 49.4 \\
(11) MVCNN (ICCV 2015) f.t., 12x (\cite{Su2016Multi}) & ImageNet1K  &ModelNet40  &12 & 12 & 70.1 \\
(12) MVCNN (ICCV 2015) f.t.+ metric, 12x (\cite{Su2016Multi}) & ImageNet1K  &ModelNet40  &12 & 12 & 80.2 \\
(13) MVCNN (ICCV 2015) 80x (\cite{Su2016Multi}) & ImageNet1K  &-  &80 & 80 & 36.8 \\
(14) MVCNN (ICCV 2015) f.t., 80x (\cite{Su2016Multi}) & ImageNet1K  &ModelNet40  &80 &80 & 70.4 \\
(15) MVCNN (ICCV 2015) f.t.+ metric, 80x (\cite{Su2016Multi}) & ImageNet1K  &ModelNet40  &80 &80 & 79.5 \\

\hline
(16) GIFT (CVPR 2016) (\cite{Zhaoxiang2016GIFT}) & ModelNet40  & - & - & - & 81.9 \\
(17) PANORAMA-NN (3DOR2017), (\cite{N20220:2017}) & ImageNet1K & ModelNet40  & - & - & 83.5 \\
(18) GVCNN (CVPR 2018), f.t. + 8x (\cite{Feng2018GVCNN})  & ImageNet1K & ModelNet40  & 8 & 8 & 79.7 \\
(19) GVCNN (CVPR 2018), f.t. + metric, 8x (\cite{Feng2018GVCNN})  & ImageNet1K & ModelNet40  & 8 & 8 & 84.5 \\
(20) GVCNN (CVPR 2018), f.t. + 12x (\cite{Feng2018GVCNN})  & ImageNet1K & ModelNet40  & 12 & 12 & 81.3 \\
(21) GVCNN (CVPR 2018), metric, 12x (\cite{Feng2018GVCNN})  & ImageNet1K & ModelNet40  & 12 & 12 & 85.7 \\
(22) CNN+LSTM (TMM 2019) (\cite{Ma2019Learning}) & ModelNet40  & - & - & - & 84.34 \\

\hline
(23) MDPCNN (Only Batch), 12x &ImageNet1K, & ModelNet40  &3 &3 & 84.2 \\
(24) MDPCNN (Batch + Sample Selection), 12x &ImageNet1K, & ModelNet40  &3 &3 & 85.5 \\
(25)\textbf{MDPCNN} (All), 12x &ImageNet1K, & ModelNet40  &3 &3 & \textbf{87.6} \\

\hline
* f.t.=fine-tuning, metric=low-rank Mahalanobis metric learning
\end{tabular}
\end{center}
\end{table*}

\subsection{Importance Analysis of Different Parts in \textbf{MDPCNN}}

In order to prove the contributions of our work, we will analyze them step by step. In detail, we will first discuss the importance of the multi-batch structure. In our experiments, the multi-batch structure is added into GPCNN where the pairwise samples are randomly selected, and only the contrastive loss is employed, and we call it as \textbf{MDPCNN} (Only Batch);  In “MDPCNN (Only Batch)” and “MDPCNN (Batch + Sample Section), only the contrastive loss is employed where the value of $\alpha$ is equal to 1 and $\beta = 0$.  From Fig.9, Fig.10, Fig.11 and Table 2, we can observe that the multi-batch structure is very important, and it is very useful to improve the performance of GPCNN. Secondly, in \textbf{MDPCNN} (Only Batch), the random sample selection is utilized, thus, in order to assess the importance of sample selection, the clustering sample selection scheme is further added in \textbf{MDPCNN} (Only Batch), and we call this method as \textbf{MDPCNN} (Batch + Sample Section) where only the contrastive loss is employed; From them, we also can see that the clustering sample selection scheme can further improve the performance of \textbf{MDPCNN} (Only Batch). The reason is that the samples that are most difficult to identify are chosen to construct the pairwise samples, and then these samples are used to train \textbf{MDPCNN}, thus, when the model can classify these most difficult samples, it will be easy for it to recognize other samples. Finally, we also discuss the role of the discriminative loss function. In detail, the contrastive-center loss is further added into \textbf{MDPCNN} (Batch + Sample Section), and we label this new scheme as \textbf{MDPCNN} where the contrastive loss and contrastive-center loss are jointly optimized. From above figures, we can find that since the category information is further employed, which can add the discriminative power of the deep learning feature, and \textbf{MDPCNN} obviously outperforms the \textbf{MDPCNN} (Batch + Sample Section). Especial for MVRED and NTU60, the improvements are very obvious.

\begin{figure}[h]
\begin{minipage}{1\linewidth}
  \centerline{\includegraphics[width=3.2in,height = 2in]{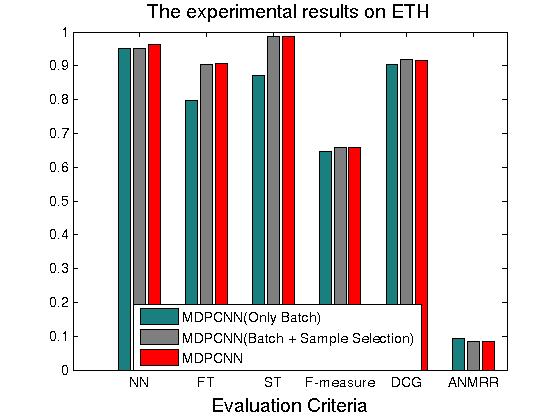}}
  \centerline{(a) Retrieval Comparison }
\end{minipage}
\hfill
\begin{minipage}{1\linewidth}
  \centerline{\includegraphics[width=3.2in,height = 2in]{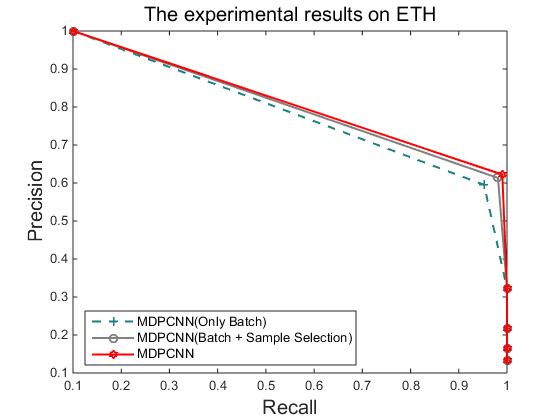}}
  \centerline{(b) PR Comparison}
\end{minipage}
\caption{Importance analysis of different parts in \textbf{MDPCNN} on ETH dataset}
\end{figure}

\begin{figure}[h]
\begin{minipage}{1\linewidth}
  \centerline{\includegraphics[width=3.2in,height = 2in]{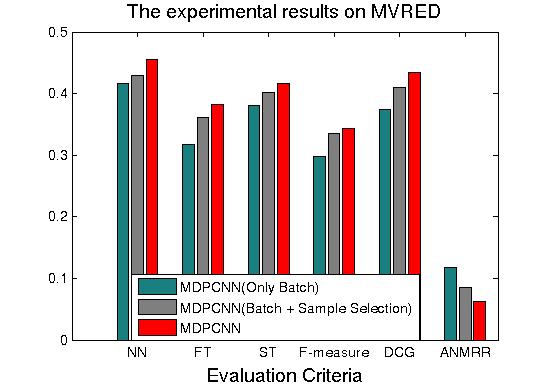}}
  \centerline{(a) Retrieval Comparison }
\end{minipage}
\hfill
\begin{minipage}{1\linewidth}
  \centerline{\includegraphics[width=3.2in,height = 2in]{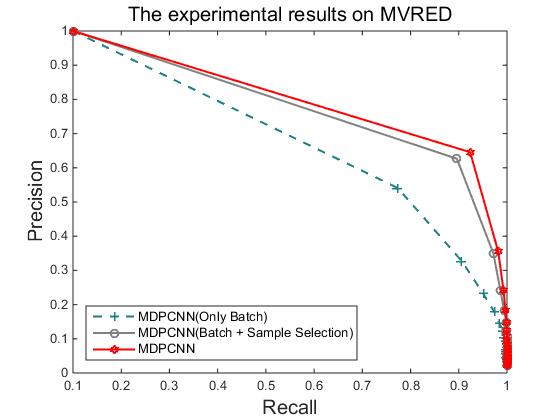}}
  \centerline{(b) PR Comparison}
\end{minipage}
\caption{Importance analysis of different parts in \textbf{MDPCNN} on MVRED dataset}
\end{figure}

\begin{figure}[h]
\begin{minipage}{1\linewidth}
  \centerline{\includegraphics[width=3.2in,height = 2in]{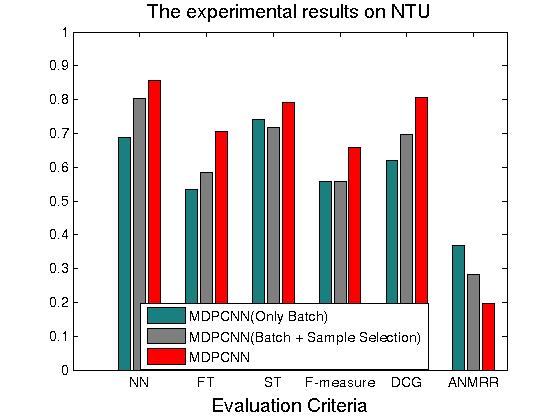}}
  \centerline{(a) Retrieval Comparison }
\end{minipage}
\hfill
\begin{minipage}{1\linewidth}
  \centerline{\includegraphics[width=3.2in,height = 2in]{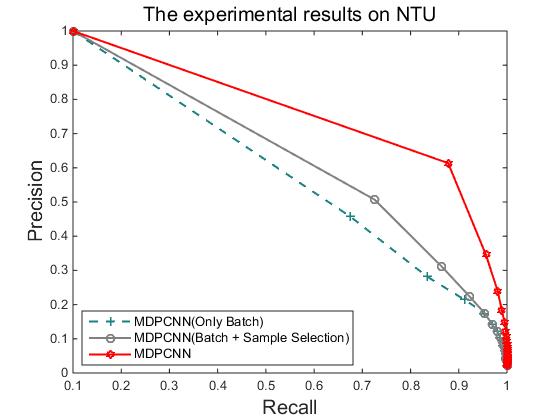}}
  \centerline{(b) PR Comparison}
\end{minipage}
\caption{Importance analysis of different parts in \textbf{MDPCNN} on NTU60 dataset}
\end{figure}

\subsection{Convergence analysis}
To further illustrate the stability of \textbf{MDPCNN}, we will show the loss convergence of the network. As shown in the Fig.12, we select 1000 pairwise samples including 500 positive sample and 500 negative samples, from ETH dataset as the training set, and then train the network with different batch sizes. Furthermore, the convergence speed of the training stage on all three datasets are given in Fig. 12 (b). From it, we can see that the network with the big batch size is much more stable. Meanwhile, the convergence speed of all three datasets is very fast, for example, ETH dataset converges at the 9000th iteration (3 epochs), and MVRED and NTU can also converge at the 15000th iteration (5 epochs). In a word, \textbf{MDPCNN} has good convergence.

\begin{figure}[h]
\begin{minipage}{1\linewidth}
  \centerline{\includegraphics[width=3.2in,height = 2in]{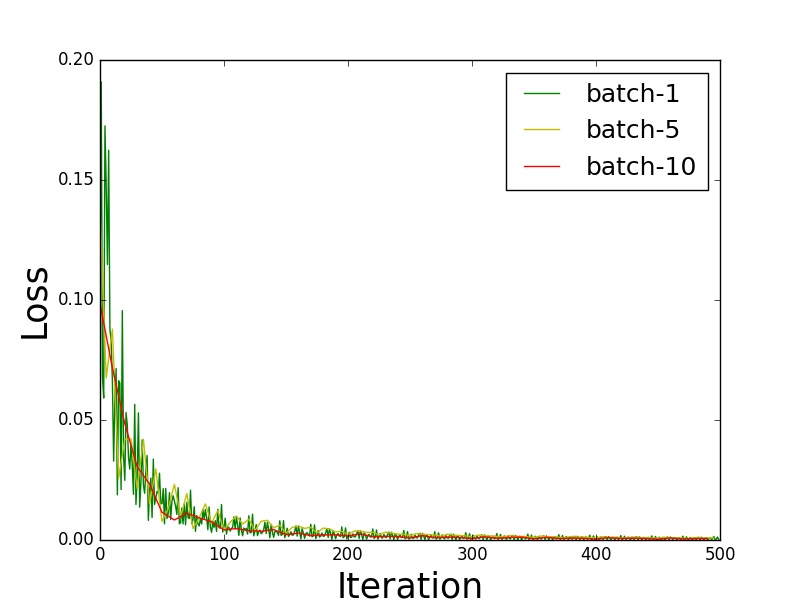}}
  \centerline{(a) Convergence stability }
\end{minipage}
\hfill
\begin{minipage}{1\linewidth}
  \centerline{\includegraphics[width=3.2in,height = 2in]{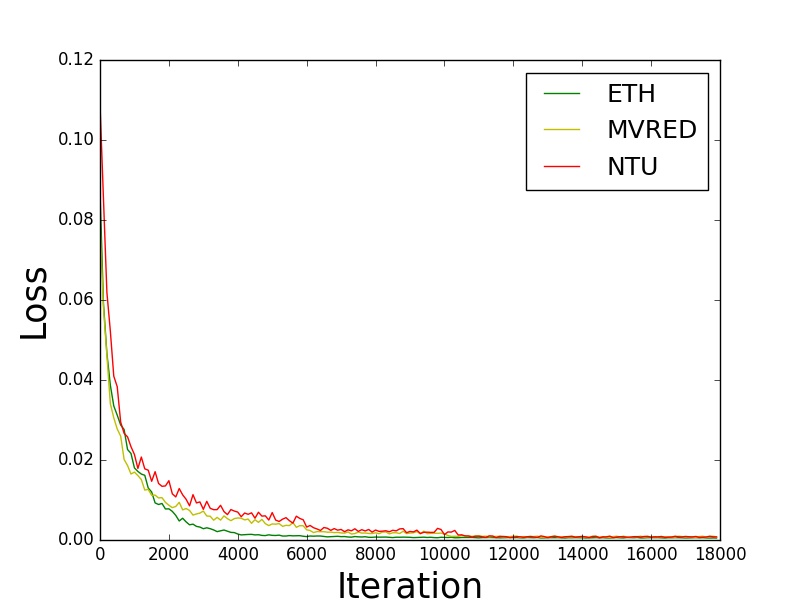}}
  \centerline{(b) Convergence speed}
\end{minipage}
\caption{ Convergence analysis, a) different batch sizes are utilized on ETH dataset, b) the convergence speed of the training process on different datasets are given.}
\end{figure}

\section{Conclusion}
In this work, we propose a novel \textbf{MDPCNN} approach for view-based 3D object retrieval task. In \textbf{MDPCNN}, multiple batches and multiple views can be simultaneously inputted by adding the Slice layer and the Concat layer. Meanwhile, a highly discriminative network can be obtained by samples selection scheme, where samples of the most difficult to identify are chosen. In addition, the features with better intra-class compactness and interclass separability are obtained by the discrimination loss function. Massive experiment results on different public 3D object datasets demonstrate that \textbf{MDPCNN} has more advantages than the state-of-the-art methods in 3D object retrieval. The ablation study also shows that when multi-batch samples are simultaneously calculated,  it can promote the improvement of \textbf{MDPCNN}. Moreover, the clustering sample selection scheme where the most difficult samples are chosen to construct the pairwise samples, is also very helpful to further improve the performance. Finally, the contrastive loss and contrastive-center loss are very effective for increasing the discriminative power of the deep learning feature.

In the near future, we will search how to process the feature map filtered by view pooling layer and how to build a more reasonable loss function.

\ifCLASSOPTIONcaptionsoff
  \newpage
\fi



%

\bibliographystyle{IEEEtran}
 \bibliography{MDPCNN19}

\end{document}